\documentclass[conference]{IEEEtran}
\usepackage{times}

\usepackage[sort,numbers]{natbib}
\usepackage{multicol}
\usepackage[bookmarks=true]{hyperref}

\usepackage{hyperref}
\usepackage{url}

\usepackage{amsmath}
\usepackage{amssymb}
\usepackage{mathtools}
\usepackage{amsthm}

\usepackage[utf8]{inputenc} 
\usepackage[T1]{fontenc}    

\usepackage{scalefnt,letltxmacro}
\LetLtxMacro{\oldtextsc}{\textsc}
\renewcommand{\textsc}[1]{\oldtextsc{\scalefont{1.10}#1}}

\usepackage{caption}
\usepackage{subcaption}
\usepackage{microtype}
\usepackage{textgreek}
\usepackage{graphicx}         
\usepackage{booktabs,tabularx}          
\usepackage{nicefrac}                   
\usepackage{graphics}                   
\usepackage{multirow}
\usepackage{lipsum}  
\usepackage[font=footnotesize]{caption}
\usepackage{algorithm}
\usepackage{algorithmic}
\usepackage{adjustbox}
\usepackage{soul}
\usepackage{xcolor}
\usepackage{pdfpages}
\usepackage{listings}
\usepackage{xspace}

\usepackage{colortbl} 
\usepackage{xcolor}   
\IEEEoverridecommandlockouts  

\newcommand{\ours}{Demo-SCORE\xspace}

\theoremstyle{plain}

\theoremstyle{definition}

\theoremstyle{remark}

\begin{document}

\title{Curating Demonstrations using Online Experience}
\author{Annie S. Chen$^*$\thanks{$^*$Equal contribution. Correspondence to \href{mailto:asc8@stanford.edu}{asc8@stanford.edu}. Videos of real-world trials at \href{https://anniesch.github.io/demo-score/}{https://anniesch.github.io/demo-score/}.}, Alec M. Lessing$^*$, Yuejiang Liu, Chelsea Finn \\
Stanford University
}

\maketitle

\begin{abstract}
Many robot demonstration datasets contain heterogeneous demonstrations of varying quality. This heterogeneity may benefit policy pre-training, but can hinder robot performance when used with a final imitation learning objective. In particular, some strategies in the data may be less reliable than others or may be underrepresented in the data, leading to poor performance when such strategies are sampled at test time. Moreover, such unreliable or underrepresented strategies can be difficult even for people to discern, and sifting through demonstration datasets is time-consuming and costly. On the other hand, policy performance when trained on such demonstrations can reflect the reliability of different strategies. We thus propose for robots to self-curate based on online robot experience (Demo-SCORE). More specifically, we train and cross-validate a classifier to discern successful policy roll-outs from unsuccessful ones and use the classifier to filter heterogeneous demonstration datasets. Our experiments in simulation and the real world show that Demo-SCORE can effectively identify suboptimal demonstrations without manual curation. Notably, Demo-SCORE achieves over 15-35\% higher absolute success rate in the resulting policy compared to the base policy trained with all original demonstrations.

\end{abstract}

\IEEEpeerreviewmaketitle

\section{Introduction}

\begin{figure}[t!]
    \centering
    \includegraphics[width=0.95\columnwidth]{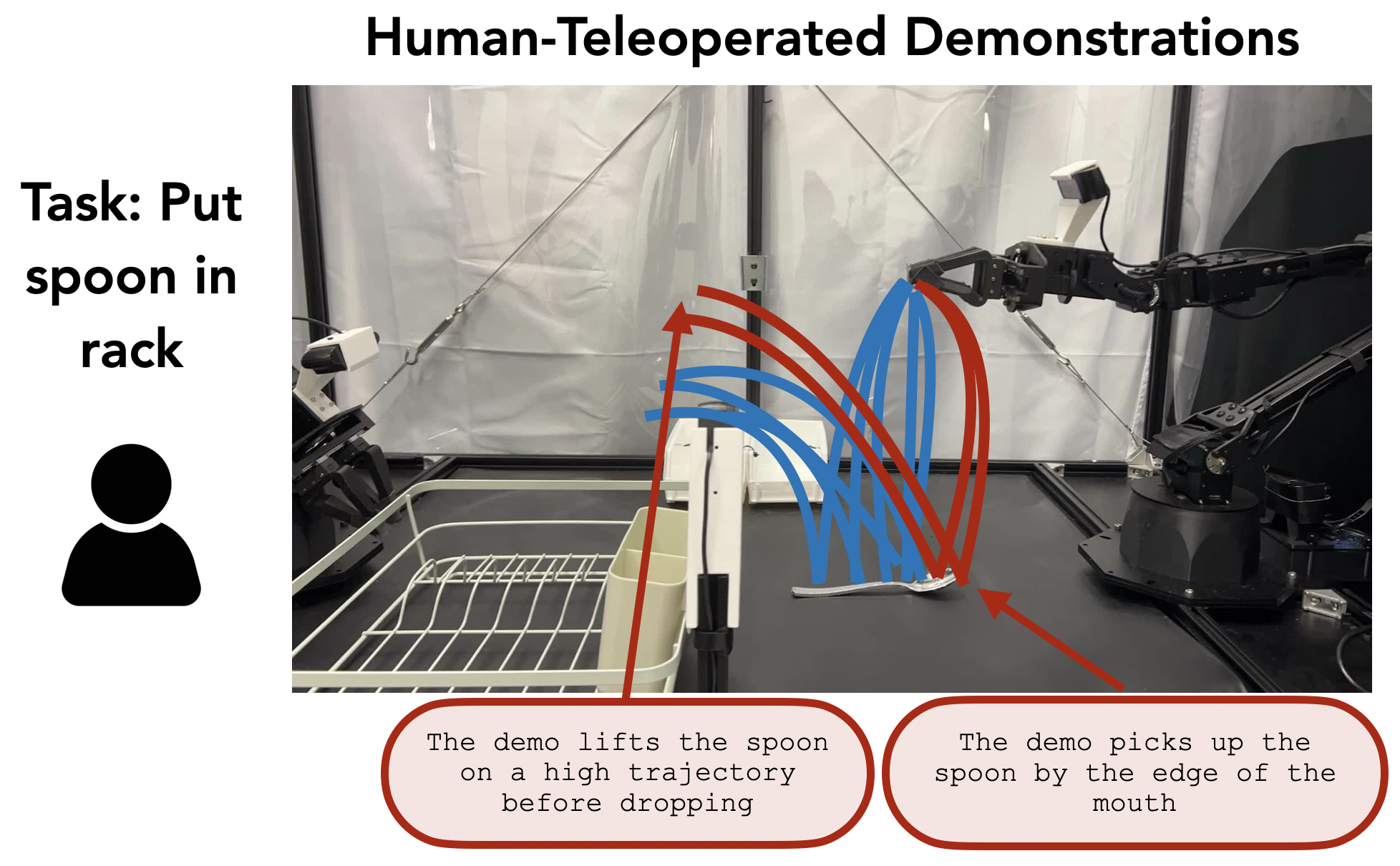}
    \caption{
        \small 
        \textbf{Human Demonstrations may be Unreliable.} 
        Human demonstrations may include a wide range of strategies, not all of which are beneficial for the robot to learn. For example, some humans may try to complete the task by picking up the spoon by its edge or by lifting it above the rack and dropping it, but these strategies may be unreliable for imitation by a robot policy.
    }
    \label{fig:teaser}
\end{figure}

Learning from demonstrations is a powerful technique for acquiring a range of different robotic manipulation skills.
Although large, diverse demonstration datasets are beneficial for pre-training, a more selective use of demonstrations may be required to achieve optimal performance.
As robot learning datasets scale to include demonstrations from multiple operators across increasingly complex tasks, they naturally encompass a wide range of strategies—some of which may be challenging for robots to replicate reliably or lack sufficient examples for effective learning. 
For example, when teaching a robot to pick up a spoon, a human operator may collect demonstrations using a strategy that may be unreliable for the robot to imitate, like gripping the edge of the spoon head, which may be prone to slipping out of the robot's grasp, as shown in Figure~\ref{fig:teaser}. 
As such, in this paper, we study how to effectively curate demonstration datasets to enable optimal policy performance. 

Manual labeling and filtering of such data is both time-consuming and error-prone, especially as the scale of robot learning datasets continues to grow. 
Moreover, determining which demonstrations are unreliable can be difficult, even for human experts, as successes for most real-world tasks exist on a spectrum of approaches with subtle variations in execution and context dependence, potentially making it difficult to describe or group different strategies together. For example, when grasping objects, demonstrations might vary continuously in terms of approach angles, contact points, and force application. 
Without careful curation, robots may learn strategies that succeed in specific demonstrations but are brittle under slightly different conditions, limiting their generalization and robustness. 
This motivates the need for an automated approach to selectively curate demonstrations \textit{within a single task/dataset}, allowing robots to learn only the most reliable and effective strategies for that task.

Our key insight is that when a policy is trained on heterogeneous demonstrations, its performance during rollouts can reveal which demonstrated strategies are most reliable. Building on this observation, we present \ours, which leverages online policy rollout experiences to automatically curate reliable demonstrations.
The approach begins by training an initial policy on the full set of demonstrations. 
Rollouts are generated from this policy and used to train a classifier that approximately distinguishes successful trajectories from failed ones, thereby learning to separate the underlying strategies by their reliability. One challenge is that classifier may overfit to rollouts from one policy checkpoint--to address this, we cross-validate its performance on rollouts from another checkpoint in the initial policy training run. 
By applying this classifier to the original dataset, \ours identifies and discards unreliable examples. 
This method works because it leverages the robot’s own experience through rollouts, allowing it to assess the true reliability of different strategies after policy learning. 
Thus, by focusing on the end results of policy rollouts, \ours filters demonstrations based on practical outcomes rather than potentially misleading surface-level cues.

We evaluate \ours on a set of simulated and real-world robotic manipulation tasks with heterogeneous demonstration mixtures that include a variety of strategies. 
More specifically, our evaluation suite spans multiple tasks in Robosuite~\cite{mandlekarWhatMattersLearning2022} along with simulated and real-world ALOHA~\cite{zhaoLearningFineGrainedBimanual2023} with different imbalanced mixtures of higher and lower quality demonstrations, and we evaluate with different policy classes, including both Diffusion Policy~\cite{chi2023diffusion}.  ACT~\cite{zhaoLearningFineGrainedBimanual2023}. 
Our experiments show that policies trained on our filtered demonstrations significantly outperform the base policies trained on all demonstrations, leading to 15-35\% higher absolute success rate. \ours also significantly outperforms prior approaches that also leverage rollout data, like autonomous imitation learning, achieving almost 2x lower failure rate. We additionally provide a detailed empirical analysis of the design choices of our method and show that it is robust to these and to hyperparameter choices. 
Our results highlight the benefits of curating demonstration data and show that our simple approach to automated filtering can significantly enhance policy performance in robot learning from demonstrations.

\section{Related Work}

{\bf Behavior Cloning.}
Behavior cloning has traditionally been formulated as supervised learning from homogeneous expert demonstrations~\citep{argallSurveyRobotLearning2009,ravichandarRecentAdvancesRobot2020,zareSurveyImitationLearning2024}. 
In practice, however, human demonstrations often exhibit wide variations in strategy~\citep{xieLearningLatentRepresentations2021}, proficiency~\citep{beliaevImitationLearningEstimating2022}, and preference~\citep{kahnUncertaintyAwareReinforcementLearning2017}, especially in large-scale datasets~\citep{mirchandaniRoboCrowdScalingRobot2024,khazatskyDROIDLargeScaleInTheWild2024,black$p_0$VisionLanguageActionFlow2024}.
To handle such variation, recent methods have sought to model the distribution of actions or action chunks~\citep{zhaoLearningFineGrainedBimanual2023,chiDiffusionPolicyVisuomotor2023,leeBehaviorGenerationLatent2024,ze3DDiffusionPolicy2024,bharadhwajRoboAgentGeneralizationEfficiency2024,wangEquivariantDiffusionPolicy2024,haldarBAKUEfficientTransformer2024,liuBidirectionalDecodingImproving2024}
or decompose demonstrations into shorter sequences corresponding to distinct high-level strategies~\citep{billardDiscoveringImitationStrategies2003,meteQueSTSelfSupervisedSkill2024,zhengPRISELLMStyleSequence2024}. 
Nonetheless, not all strategies are equally optimal: some can be inefficient, overly complex, or brittle to perturbations, thereby hindering robust deployment~\citep{wuImitationLearningImperfect2019,brownExtrapolatingSuboptimalDemonstrations2019,palejaHeterogeneousLearningDemonstration2020,jayanthiDROIDLearningOffline2023,kuharLearningDiscernImitating2023}. Our method addresses this issue by explicitly curating demonstration datasets.


{\bf Data Curation.} 
Early efforts in data curation often rely on hand-crafted heuristics, such as near-duplicate removal~\citep{lee2022deduplicating}, metadata balancing~\citep{van-esch-etal-2022-writing}, and diversity expansion~\citep{gaoEfficientDataCollection2024}.
While manual curation has shown promise in the development of robot foundation models~\citep{black$p_0$VisionLanguageActionFlow2024}, it is inherently limited in its ability to capture nuanced aspects of data quality.
For example, the commonly used heuristic of maximizing state diversity has been shown not to always be beneficial~\citep{belkhaleDataQualityImitation2023}.
Recent works have instead shifted towards curating training data based on properties derived from the learned policy.
Notably,~\citet{duBehaviorRetrievalFewShot2023a,nasirianyLearningRetrievalPrior2023} leverage the embeddings of a small set of high-quality demonstrations to retrieve relevant examples from a larger candidate dataset.
~\citet{hejnaReMixOptimizingData2024} propose to estimate the importance weights of subsets through the lens of domain robustness~\citep{xieDoReMiOptimizingData2023}.
However, estimating the quality of individual demonstrations within a single task remains an open challenge.
Our work addresses this challenge by identifying and pruning suboptimal demonstrations.

{\bf Suboptimal Demonstrations.}
Previous works have attempted to tackle suboptimal demonstrations in two primary ways.
One line of research assumes access to annotations, such as reward~\citep{tangkarattVariationalImitationLearning2020} or ranking~\citep{brownExtrapolatingSuboptimalDemonstrations2019}, for all collected demonstrations, aiming to learn an optimal policy via weighted behavior cloning~\citep{xuDiscriminatorWeightedOfflineImitation2022a,wangLearningWeightImperfect2021} or offline reinforcement learning~\citep{levineOfflineReinforcementLearning2020}.
However, gathering dense annotations at scale beyond sparse binary success detection can be tedious and expensive.
As such, another line of work seeks to learn a reward function from a limited number of annotations (e.g., quality~\citep{xuDiscriminatorWeightedOfflineImitation2022a}, confidence~\citep{sasakiBehavioralCloningNoisy2020}, preference~\citep{caoLimitedPreferenceAided2024}) and subsequently estimate the rewards of a broader unlabeled dataset.
While these methods reduce annotation expenses, they still rely on {\em human definitions of data quality}, which does not always align with robot capabilities or task conditions.
In contrast, we leverage policy rollouts to identify suboptimal trajectories autonomously and only use sparse binary success/failure annotations of the data. 
Prior works like autonomous imitation learning~\cite{bousmalis2023robocat,ahn2024autort,mirchandani2024so} and reward-conditioned policies~\cite{kumar2019reward,schmidhuber2019reinforcement} also leverage policy rollouts filtered by success as additional data to improve the policy. Compared to the these, we find in Section~\ref{sec:exps} that \ours is more effective at improving policy performance. 
By grounding curation directly in the downstream task and embodiment, our method provides a scalable, effective pathway for filtering demonstrations.


\section{Preliminaries}

We address a problem setting where we are given a dataset of $N$ successful demonstrations, $\mathcal{D}_{\text{demo}} = \{\tau_1, \ldots, \tau_N\}$. Each demonstration $\tau_i = \{s_0, a_0, s_1, a_1, \ldots, s_T, a_T\}$ consists of a trajectory of states and actions. These demonstrations are collected in a Markov Decision Process (MDP) characterized by $(\mathcal{S}, \mathcal{A}, P, r)$, where $\mathcal{S}$ and $\mathcal{A}$ denote the state and action spaces, $P$ is the transition dynamics, $r$ is the reward function. The goal is to learn a policy $\pi_\theta(a | s)$ that maximizes expected cumulative reward, $\mathbb{E}[R] = \mathbb{E}[\sum_{t=0}^T \gamma^t r(s_t, a_t)]$, during deployment. We additionally have the opportunity to collect a small amount of experience in the environment for further offline training, where experience is labeled with a binary indicator for success or failure.

We assume that all of the demonstrations in $\mathcal{D}_{\text{demo}}$ are successful (i.e. achieve reward $r=1$), or equivalently that any unsuccessful demonstrations are thrown out. Beyond being successful, motivated by the discussion above, we focus on settings where the demonstrations are heterogenous and differ in reliability and effectiveness. This diversity arises due to variations in operator styles, environmental conditions, or task execution strategies. As a result, some demonstrations may lead to suboptimal behaviors if modeled indiscriminately, posing a challenge for imitation learning methods that aim to model the distribution of demonstrations. Our goal is to learn a maximally successful policy, $\pi_{\text{final}}$, despite varying demonstration quality. 

Our problem setting is a case of the autonomous imitation learning (Auto-IL)
problem.
In autonomous imitation learning~\cite{bousmalis2023robocat,ahn2024autort,mirchandani2024so}, an initial policy $\pi_0$ is trained, which is used to generate rollouts. These are processed through a filtering mechanism, often binary task success/failure, to generate a supplementary dataset. A new policy $\pi_i$ is then trained or fine-tuned using a combination of demos and newly acquired data.
The online experience is limited to far fewer trajectories than the typical budget of reinforcement learning methods, making it particularly practical for real robots. In the next section, we will describe how we use this experience to identify how to filter the demonstrations to improve performance.

\section{\ours Data Curation}

\begin{figure*}[h!]
    \centering
    \includegraphics[width=1.0\textwidth]{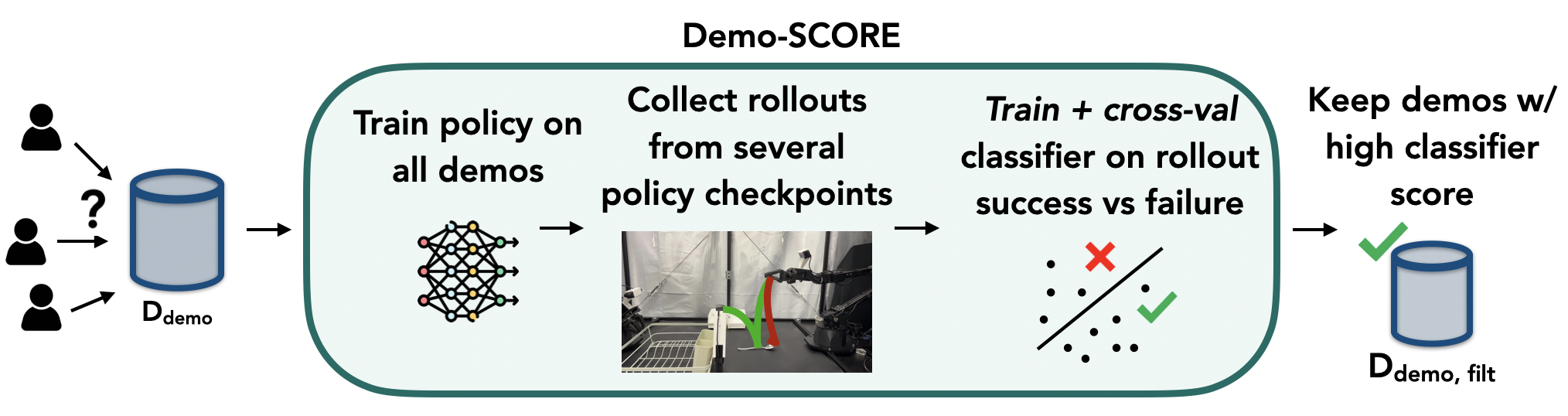}
    \caption{
        \small 
        \textbf{Illustration of \ours.} Our method self-curates demonstrations through online robot experience in four steps: (i) train a policy on the full demonstration set; (ii) evaluate the policy by generating rollouts with different checkpoints; (iii) use these rollouts to train and cross-validate a classifier that distinguishes between success and failure trajectories; (iv) filter out unreliable demonstrations based on the learned classifier, retaining only ones that do not reflect policy failures for subsequent policy training.
    }
    \vspace{-4mm}
    \label{fig:method}
\end{figure*}

In this section, we present \ours, where we seek to filter out unreliable demonstrations.
The key empirical observation behind \ours is that successes and failures in policy rollouts can reflect unreliable strategies in the training data that may not be easily discernible from the demonstrations themselves, which are all labeled as successful.
By focusing on the outcomes of policy rollouts, \ours can identify and filter out unreliable demonstrations that the robot policy cannot replicate successfully and therefore lead to inconsistent policy performance.

 Our approach consists of the following main steps: first, we train an initial policy on the full set of demonstrations. We then use the policy to generate rollouts that are used to train a data quality classifier to distinguish between successful and failed outcomes. This classifier is subsequently applied to the original dataset to perform filtering (which can be done at the episode level or the chunk level), retaining only the reliable episodes. 

Our method does not require any additional human labeling beyond the detection of success or failure during the rollout phase. This lack of manual dense supervision makes \ours scalable to large, diverse demonstration datasets where traditional human annotation would be challenging and costly.
Furthermore, grounding demo curation in policy rollouts allows us to identify suboptimal demos that are difficult for human annotators to recognize.

\subsection{Initial policy and data quality classifier training}
We first train an initial policy, $\pi_{0,K}$, on the full set of given demonstrations, $\mathcal{D}_{\text{demo}}$, for a total of $K$ steps. 
In this step, we prioritize learning a policy capable of modeling the full distribution of demonstrated behaviors while avoiding mode collapse when presented with heterogeneous demonstrations. Recently proposed policy classes such as diffusion policy~\cite{chi2023diffusion} and ACT~\cite{zhao2023learning} have demonstrated robust performance in this context, making them suitable choices for our base policy. This policy is trained until convergence, so that the model learns to replicate the different behaviors shown in the demonstrations.



After the initial training phase, we generate rollouts produced by executing the policy in the environment, which may be rollouts obtained in the process of evaluating the initial policy at various checkpoints. 
We want to use these rollouts to train a data quality classifier that distinguishes between successful and failed outcomes and therefore hopefully between reliable and unreliable strategies.
More formally, let $+$ denote reliable demonstrations and $-$ denote unreliable ones. The learned classifier is expected to provide plausible estimates of demo quality if
$p_{\text{rollout}}(\tau|\text{success}) 
\approx p_{\text{demo}}(\tau|+)$ and $p_{\text{rollout}}(\tau|\text{failure}) \approx p_{\text{demo}}(\tau|-)$, but in practice, we do not have such rollout distributions nor quality labels $+/-$. 
Thus, one key challenge is that the classifier may overfit to the rollouts it is trained on, which may not perfectly match the original demo distribution.
In particular, at convergence time, even though the policy is trained to imitate the full demo distribution and good strategies should be much more reliable than bad ones, 
the policy rollouts may still contain failures near good strategies and successes near bad strategies. 
Overfitting to a rollout distribution from one policy checkpoint could lead to poor generalization on the demo distribution.


To address this challenge of distribution shift and potential classifier overfitting, we leverage multiple rollout distributions taken from different checkpoints across the initial training run and use cross-validation. We obtain rollouts at $C$ evenly spaced checkpoints of the initial training run, giving $\{\mathcal{D}_{\pi_{0, i}}\}_{i=1}^C$, each taken at checkpoint $i*K/C$. 
We then train $C-1$ classifiers, $\{q_{\phi_i}\}_{i=1}^{C-1}$, on rollouts from each of the first $C-1$ checkpoints, and perform cross-validation with rollouts from the last checkpoint $\mathcal{D}_{\pi_{0, C}}$ as the validation set. 
Intuitively, the rollout distribution changes throughout the course of training, typically being broader at the beginning and becoming narrower as the policy fits the original demonstrations. We choose to use $C$ spaced-out checkpoints for several reasons. First, they are spread apart enough to ensure that the rollout distributions are distinct from each other, so a classifier that overfits to rollouts from a checkpoint would be less likely to be chosen through cross-validation. Second, when evaluating the initial training run, it is common to evaluate performance at several evenly spaced checkpoints, so these online rollouts occur little additional online cost beyond evaluating the original training run.
Other choices of checkpoints are also effective, as we find in our testing of variations of \ours in Section~\ref{sec:exps-ablations}. Furthermore, using as few as $10$ rollouts per checkpoint can be sufficient, for a total of less than 100 rollouts.

Each data quality classifier $q_{\phi_i}$ is trained to predict whether a state from the corresponding set of rollouts $\mathcal{D}_{\pi_{0, i}}$ is from a successful or a failed trajectory.
The classifiers are trained using a binary cross-entropy loss, where the input consists of the proprioceptive state $s_t$, and the output is a binary label $y_i \in \{0, 1\}$ indicating the outcome of the corresponding rollout. 
For a single trajectory $\tau = \{(s_t, a_t)\}_{t=1}^T \in \mathcal{D}_{\pi_{0, i}}$ with label $y$, the loss at the trajectory level is defined as:  
\[
L_\phi(\tau, y) = \frac{1}{T} \sum_{t=1}^T \left[ -y \log q_{\phi_i}(y \mid s_t) - (1 - y) \log (1 - q_{\phi_i}(y \mid s_t)) \right],
\]
where $T$ is the number of timesteps in the trajectory $\tau$.  

The overall loss is then the average over all trajectories in the rollout dataset $\mathcal{D}_{\pi_{0, i}}$: 
\[
\mathcal{L}_\phi = \frac{1}{M} \sum_{(\tau_j, y_j) \in \mathcal{D}_{\pi_{0, i}}} L_\phi(\tau_j, y_j),
\]
where $M$ is the number of rollouts in the training dataset, $\mathcal{D}_{\pi_{0, i}}$.  

The classifiers are small MLPs (with two hidden layers with eight hidden units).
We train the classifiers with large weight decay and dropout to prevent overfitting to the rollout distribution, although we find in Section~\ref{sec:exps-ablations} that the classifier is robust even without heavy regularization due to cross-validation.
The classifier checkpoint with the lowest validation loss is chosen as the final data quality classifier, $q_{\phi^*}$  and is then used in the next phase to filter the demonstrations.

\subsection{Demonstration Filtering}
In the demonstration filtering phase, we apply the learned data quality classifier to the original dataset of demonstrations, in the following manner: Each state $s_t$ from the demonstration is passed as input to the classifier, which returns the probability of success. Averaged across all states in the episode, if the probability of success exceeds a threshold, $\gamma$,
the demonstration is kept; otherwise it is discarded. 
For an episode \(\tau = \{(s_t, a_t)\}_{t=1}^{T}\), the average success probability is  
\[
\bar{q}_{{\phi^*}}(\tau) = \frac{1}{T} \sum_{t=1}^{T_i} q_{\phi^*}(y = 1 \mid s_t).
\]
We use the average probability of success over all states in the rollout training dataset $\mathcal{D}_{\pi_{0, i^*}}$, corresponding to the training set of the best classifier $q_{\phi^*}$, as the threshold, e.g. 
\[ \gamma = \frac{1}{|\mathcal{D}_{\pi_{0, i^*}}|} \sum_{(s_t) \in \mathcal{D}_{\pi_{0, i^*}}} q_{\phi^*}(y =1 \mid s_t),\]
and the filtered dataset is then:  
\[
\mathcal{D}_{\text{demo, filt}} = \{ \tau_j \mid \bar{q}_{{\phi^*}}(\tau_j) > \gamma, \tau_j \in \mathcal{D}_{\text{demo}} \}.
\]

$\gamma$ can be tuned further but we find this formulation to work well empirically for all of our experiments in Section~\ref{sec:exps}. 
The probability learned by the classifier of the failures should be much lower than those of successes, which is why $\gamma$ is an effective threshold, even if the quality composition of the dataset is imbalanced, and we find this to be the case in our experiments.


We can then use the filtered dataset $\mathcal{D}_{\text{demo, filt}}$ to train a new policy, $\pi_{\text{final}}$, either from scratch or by fine-tuning the initial policy which was trained on the full demonstration dataset.
We then apply our same trained classifier on the successful rollouts themselves, obtaining $\mathcal{D}_{\text{rollouts, filt}}$, and we can then train on the union of $\mathcal{D}_{\text{demo, filt}}$ and $\mathcal{D}_{\text{rollouts, filt}}$ in this last stage. 
By filtering out unreliable demonstrations, \ours helps ensure that the policy learns the most effective strategies, ultimately improving the performance and robustness of the learned robot behavior. Our full method is summarized in Algorithm~\ref{algoblock1}.

\begin{algorithm}[h]
\caption{\ours Summary}
\label{algoblock1}
\begin{algorithmic}[1]  
\STATE \textbf{Input:} $\mathcal{D}_{\text{demo}}, C \text{ checkpoints} , M \text{ rollouts}, K \text{ steps}$
\STATE $\pi_0 \gets \text{Train}(\mathcal{D}_{\text{demo}})$
\STATE $\{\mathcal{D}_{\pi_{0, i}}\}_{i = 1}^{C}, \gets \text{GenerateRollouts}(\pi_{0, i*K/C}, M)$
\STATE $q_{\phi^*} \gets \text{TrainClassifier, CrossValidate}(\{\mathcal{D}_{\pi_{0, i}}\}_{i = 1}^{C})$
\STATE $\gamma \gets \frac{1}{|\mathcal{D}_{\pi_{0, i^*}}|} \sum_{(s_t) \in \mathcal{D}_{\pi_{0, i^*}}} q_{\phi^*}(y =1 \mid s_t)$
\STATE $\mathcal{D}_{\text{demo, filt}}, \mathcal{D}_{\text{rollouts, filt}} \gets \{ \tau_j \mid \bar{q}_{{\phi^*}}(\tau_j) > \gamma, \tau_j \in \mathcal{D}_{\text{demo}} \cup \mathcal{D}_{\text{rollouts}}\}$
\STATE $\pi_{\text{final}} \gets$ Use $\mathcal{D}_\text{demo, filt} \cup \mathcal{D}_{\text{rollouts, filt}}$ to \text{re-train \& fine-tune} $\pi$
\STATE Return $\pi_{\text{final}}$
\end{algorithmic}
\end{algorithm}

\section{Experimental Results}
\label{sec:exps}

In this section, we aim to answer the following empirical questions: 
\begin{enumerate}
    \item Across a range of tasks and heterogenous demonstration mixtures, does \ours filter out unreliable demonstrations in a manner that leads to higher policy performance than training on all original demonstrations? 
    \item How does \ours compare to prior methods that leverage online rollouts to improve policy performance?
    \item What can the composition of demonstrations filtered out by \ours inform us about the quality of the original dataset?
    \item How sensitive is \ours to method design choices and hyperparameters?
\end{enumerate}

To answer the first two questions, we describe experiments for two simulated settings and then present results from several real-world robotic manipulation tasks with ALOHA.
We then analyze our method through ablation studies to answer the last two questions. 

\subsection{Experimental Setup}

\paragraph{General Experimental Setup}
We use base learning algorithms Diffusion Policy~\cite{chi2023diffusion} for all experiments in Robosuite
and ACT~\cite{zhaoLearningFineGrainedBimanual2023} for simulated and real-world ALOHA experiments. 
Both of these policy classes have been shown to fit the distribution of heterogeneous demonstrations.
We implement all methods on top of state-of-the-art implementations of these policies using the original diffusion policy codebase, original ACT codebase for real-world experiments, and the LeRobot codebase~\cite{cadene2024lerobot} for simulated ALOHA experiments.


\paragraph{Comparisons}
We evaluate \ours along with the following prior methods, the last two of which also leverage online rollouts to improve policy performance: 
(1) Policy trained on all original demonstrations, which we refer to as \textbf{Base Policy};
(2) \textbf{Training loss}~\cite{shen2019learning}, which weights demonstrations inversely proportional to the loss of the policy on the demonstration data.
(3) Autonomous imitation learning (\textbf{Auto-IL})~\cite{bousmalis2023robocat,ahn2024autort,mirchandani2024so}, where we train the policy on all original demonstrations and successful online rollout episodes; 
(4) Return-conditioned policies (\textbf{RCP})~\cite{kumar2019reward, schmidhuber2019reinforcement}, which trains a policy with an additional input associated with return (here 1 if from a success trajectory, 0 for failure) of the state-action pair on all demos and rollout data; 
For our main simulated experiments, we use 100 rollouts from $C=4$ checkpoints for \ours, and all 400 rollouts for Auto-IL and RPC baselines. For real-world experiments, we use 20 rollouts per checkpoint for Chocolate and Strawberry tasks and 25 for the Jellybean, Sharpie, and Spoon tasks for $C=5$ checkpoints to train the classifier. For evaluation, we measure and report success rate and 90\% confidence intervals across 256 rollouts in simulated experiments and 30 rollouts in real.

\begin{figure*}[t!]
    \centering
    \includegraphics[width=1.0\textwidth]{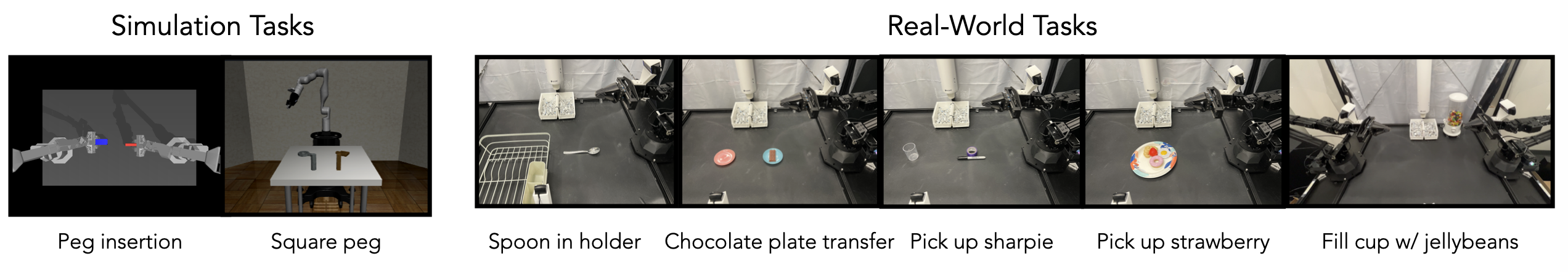}
    \caption{
        \small 
        \textbf{Tasks for Evaluation.} In simulation, we evaluate \ours on a bimanual peg insertion task in an ALOHA environment and a square peg task from Robosuite. 
        In the real-world, we evaluate on four tasks with ALOHA: spoon in rack cupholder, chocolate plate transfer, sharpie pick up, and strawberry pick up from a cluttered scene.
    }
    \label{fig:tasks}
\end{figure*}


\subsection{Simulated Experiments in Robosuite and ALOHA}

We test \ours on a bimanual peg insertion task in the ALOHA~\cite{zhaoLearningFineGrainedBimanual2023} environment and a square peg task from Robosuite~\cite{mandlekarWhatMattersLearning2022}, shown in Figure~\ref{fig:tasks}. In the peg insertion task, an ALOHA simulated in MuJoCo picks up a peg and a socket in each hand and inserts the peg into the socket. The observation space for this task consists of joint positions and an overhead RGB camera. In the square peg task, a simulated Panda arm is tasked with picking up a large square-shaped nut with a handle attached and inserting or dropping it on square-shaped peg fixed to the table. Observations for this task consist of proprioception for the robot arm and pose for the square nut.

\paragraph{Demonstration Collection}
For the peg insertion task, we collect demonstrations by scripting three policies (PegA, PegB, PegC) and saving rollouts in which the task was successfully completed. In demos from PegA, the grippers pick up the peg and socket near the center of gravity of both objects with both arms at natural positions. Without changing the orientation of the peg and socket, the arms align the peg and socket and then come together to insert them. 
In demos from PegB and PegC, when picking up the peg and socket, the positioning of the arms may sometimes obscure the overhead camera's view of the peg and socket during insertion, which may lead to worse policy performance with this strategy.

For the square peg task, we leverage the proficient human-collected dataset given in the benchmark, denoted Human, collected by one human skilled at the task. Additionally, we add demonstrations from two scripted policies: 
SquareA, in which the gripper grasps the midpoint on the side of the square, moves the nut over the peg, and then drops the nut, SquareB, in which the gripper grasps the nut across the diagonal at the corner of the square and then drops the nut on the peg. 

Although they use different strategies, all of the demonstrations successfully complete the task and we do not add any artificial noise. 
We evaluate our method on a variety of heterogeneous data mixtures taken from a combination of the demonstration sources. In particular, these mixtures include both even and strongly lopsided ratios of different demonstration strategies, in order to represent the varied potential heterogeneity in demonstration datasets.

\paragraph{Results}

\begin{figure*}[t!]
    \centering
    \includegraphics[width=1.0\textwidth]{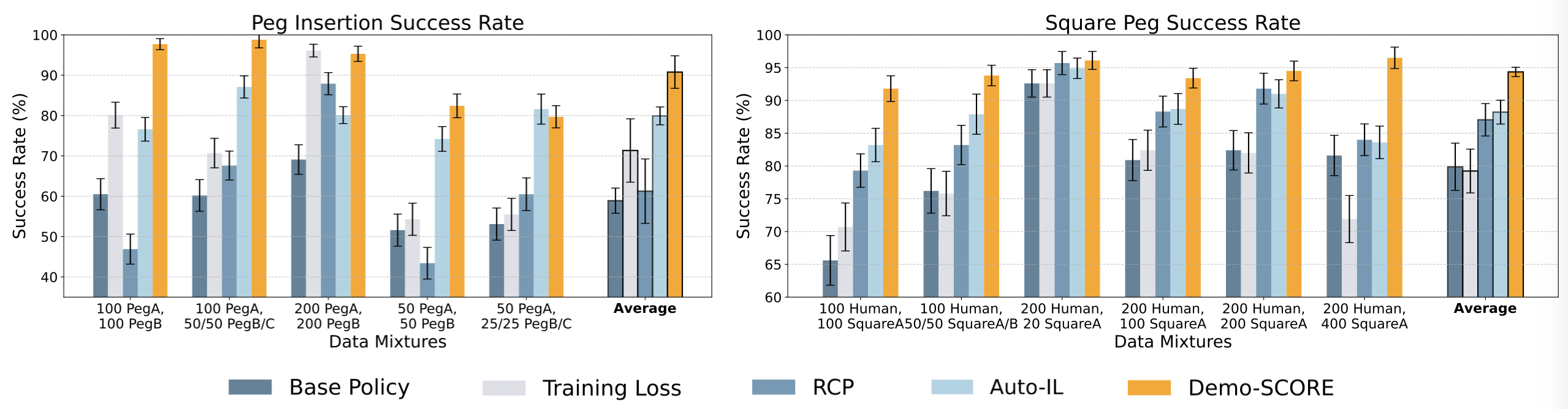}
    \caption{
        \small 
        \textbf{Simulated Experiment Results.} We report the success rates and 90\% confidence intervals of the policy trained on the filtered demonstrations by \ours along with prior methods leveraging online rollouts, across 256 trials for a variety of heterogeneous demonstration data mixtures. For example, in the square peg domain, ``100 Human, 50/50 SquareA/B'' indicates a demonstration dataset with 100 trajectories  from the proficient human dataset, 50 from the SquareA script, and 50 from SquareB. Across both tasks and all data mixtures, \ours matches or outperforms all other methods.
    }
    \label{fig:simexps}
\end{figure*}

In Figure~\ref{fig:simexps}, we show the results of \ours on the simulated peg insertion and square peg tasks for a variety of data mixtures.
We find that on average over all data mixtures in both settings, \ours improves upon training on all original demonstrations by 15-35\% absolute success rate.
With filtering by \ours, the policy achieves an average success rate of over 94\% on the square peg and over 90\% on the peg insertion task across all mixtures.
Policies trained with \ours have average failure rates that are on average over 2$\times$ lower than those of the best prior method (Auto-IL) for the square task and 1.6 times lower for the peg insertion task.
Auto-IL is the second-best method, achieving 88.2\% average success rate on the square task and 79.9\% on the peg insertion task.
It improves upon the base policy, as it is trained with more data from successful rollouts, but unreliable demonstrations are still present in the dataset, which still leads to inconsistent policy performance.
The Training Loss comparison improves upon training on all original demonstrations for the peg insertion task but not for square peg, suggesting that unreliable strategies may not necessarily correspond to ones that are easiest or fastest to learn. 
Return-conditioned policies outperforms the base policy on both tasks but performs significantly worse than \ours, suggesting that training with the return signal is not as effective in pushing the policy away from unreliable strategies as better curating demonstrations.

Qualitatively, \ours filters out between 16\% and 67\% of demonstrations for the square peg task, depending on the data mixture. We note \ours is effective even with strongly lopsided heterogeneous mixtures, such as 200 Human, 20 SquareA or 200 Human, 400 SquareA, showing that \ours adjusts accordingly based on the dataset composition.

\subsection{Real-world Experiments}

We test \ours on several tasks requiring precise manipulation with a real-world ALOHA~\cite{zhaoLearningFineGrainedBimanual2023} environment, shown in Figure~\ref{fig:tasks}:
(1) Spoon, where the robot needs to pick up a spoon and put it in the cupholder on a rack;
(2) Chocolate, where the robot needs to transfer a chocolate bar from one plate to another;
(3) Sharpie, where the robot needs to pick up the sharpie and avoid the adjacent tape roll;
(4) Strawberry, where the robot needs to pick up a strawberry from a cluttered breakfast platter; 
(5) Jellybean, where the robot must take a cup from a dispenser, place it under the jellybean holder, add jellybeans to the cup, and move the filled cup to the front of the table. The base policy trained on all demonstrations for the Jellynean task cannot complete the full task. Therefore, for this task alone, we break the task into five subtasks and use an aggregated subtask score — the fraction of the five subtasks completed in a given episode — for evaluation and classifier training supervision. 

\paragraph{Demonstration Collection}
We collect a total of 40 successful demonstrations for the Spoon task, 50 for the Chocolate and Sharpie tasks, and 120 for the Strawberry task.
We aim to reflect the diversity of demonstrations found in crowdsourced demonstrations, including a mix of different strategies.
In particular, for each of these tasks, we collect demonstrations picking up the object from different grasp positions and angles, encompassing a variety of initial conditions for the objects.
For the Jellybean task, we use 124 crowdsourced demonstrations and 100 expert demonstrations from \citet{mirchandaniRoboCrowdScalingRobot2024}.


\paragraph{Results}

\begin{figure*}[t!]
    \centering
    \includegraphics[width=1.0\textwidth]{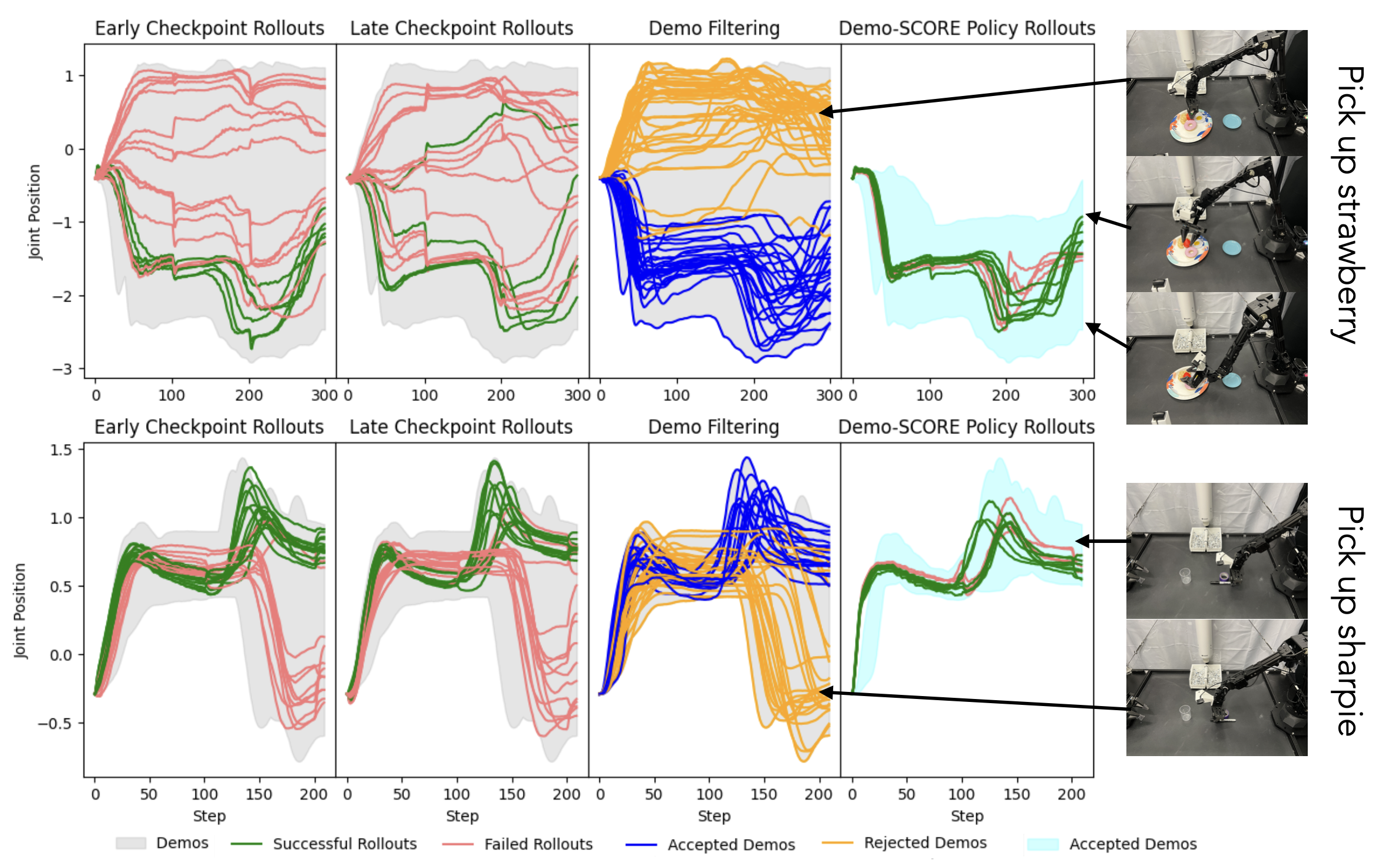}
    \caption{
        \small 
        \textbf{Rollout and Demo distributions on real-world tasks.} For the strawberry and sharpie tasks, we show the rollout distributions at early and late checkpoints of the initial policy run,
        the composition of demonstrations kept by \ours compared to the original distribution of demonstrations, and the resulting rollout distribution after filtering.
        The x axis represents step number in the rollout, and the y axis represents the first PCA component of the joint positions. 
    }
    \label{fig:rollout_demos}
\end{figure*}

\begin{figure}[t]
    \centering
    \includegraphics[width=1.0\columnwidth]{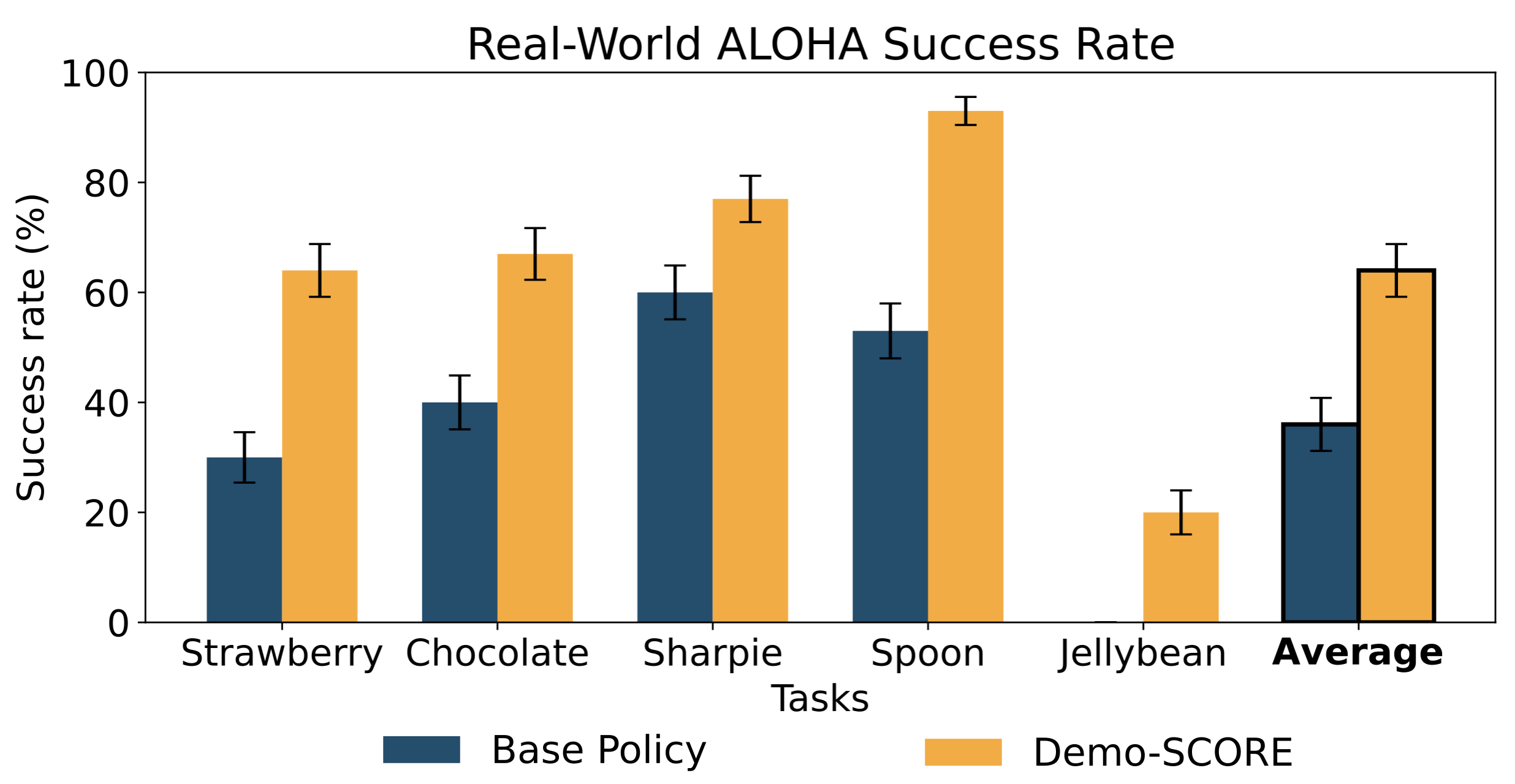}
    \caption{
        \small 
        \textbf{Real-World Results with ALOHA.} We report the success rate and 90\% confidence interval of the policy trained on the filtered demonstrations by \ours compared to training on all original demonstrations across 30 trials.
        \ours improves policy performance almost 30\% absolute success rate on average across four real-world tasks.
    }
    \label{fig:realexps}
\end{figure}

\begin{table}
\centering
\begin{tabular}{lcc}
\toprule
                          & Base Policy & \ours    \\
                          \toprule
Retrieved Cup             & 70            & 96                \\
Placed Cup                & 50            & 64                \\
Moved Cup Below Dispenser & 10            & 44                \\
Filled Cup                & 0             & 20                \\
Delivered Filled Cup      & 0             & 20                \\
\midrule
Aggregated Score          & 27            & \textbf{49}  \\  
Full Success              & 0             & \textbf{20} \\
\bottomrule
\end{tabular}
\caption{
        \small 
        We report the percent success rate for each subtask in the Jellybean task, the aggregated score (the percentage of all subtasks completed), and the percent success rate on the full Jellybean task. On this challenging task, Demo-SCORE is able to curate the data so that the robot is able to even achieve success on the full task.
    }
    \label{tab:jelly-subtask}
\end{table}

\begin{table*}[t!]
\centering
{
\begin{tabular}{@{}p{4.5cm} >{\centering\arraybackslash}p{1.5cm} >{\centering\arraybackslash} >{\centering\arraybackslash}p{2.0cm} >{\centering\arraybackslash}p{1.5cm} >{\centering\arraybackslash}p{1.5cm} >{\centering\arraybackslash} >{\centering\arraybackslash}p{1.5cm} >{\centering\arraybackslash}p{1.5cm}|>{\centering\arraybackslash}p{1.0cm}@{}}
\toprule
                                                       & \mbox{100 Human}, 100 SquareA & {\mbox{100 Human}, 50/50 SquareA/B} & {\mbox{200 Human}, 20 SquareA} & {\mbox{200 Human}, 100 SquareA} & {\mbox{200 Human}, 200 SquareA} & {\mbox{200 Human}, 400 SquareA} & {Average} \\ \toprule
\textbf{Demo-SCORE (original)}    & 91.8        & 93.8          & 96.1          & 93.4          & 94.5          & {96.5} & 94.4                      \\
Demo-SCORE (chunk)               & 90.2      & 91.4       & 94.1         & 96.1         & 95.7         & 93.0           & 93.4 \\
Demo-SCORE (trajectory)          & {93.0} & {95.7} & {97.3} &{97.3} & {96.1} & 95.3          & {95.8}             \\
Demo-SCORE (plateau checkpoints) & 92.6        & 91.0            & 96.9          & 95.3          & 94.9          & 96.1          & 94.5                      \\
Demo-SCORE (no reg)              & 94.1        & 91.4          & 92.2          & 92.2          & {96.1}          & 95.7          & 93.6                     \\
\midrule
Demo-SCORE (no cross-val)        & 93.8        & 89.1          & 87.1          & 84.4          & 81.3          & 95.3          & 88.5                      \\
\bottomrule
\end{tabular}
}
\caption{
        \small 
        \textbf{Variations of \ours.} We report the success rate of \ours with different variations in the Robomimic square peg setting.
        \ours achieves strong performance when filtering at the chunk level instead of episode level, showing that it can be applied at different levels of granularity.
        \ours also performs well using a trajectory classifier instead of a single-step classifier.
        Using checkpoints near the performance plateau of the initial training run instead of evenly spaced checkpoints is also a good option for training an effective data quality classifier.
        Finally, we find that cross-validation and regularization are crucial for \ours to prevent overfitting to the rollout distribution. 
    }
    \label{tab:ablations}
    \vspace{-3mm}
\end{table*}


In Figure~\ref{fig:realexps}, we show the results of \ours on four real-world ALOHA tasks. 
Across all four tasks, we find that \ours significantly improves upon the base policy, yielding 15-40\% absolute improvement in success rate, and almost 30\% on average. For the jellybean task, in addition to the aggregated subtask score increasing from 27 to 49 percent, the success rate on the long-horizon full task increases from 0 to 20 percent (details found in Table \ref{tab:jelly-subtask}).
These results underscore the benefits of carefully considering the quality of human-collected demonstrations--fewer demonstrations may learn a more robust policy than training on all available demonstrations if curated effectively.

In Figure~\ref{fig:rollout_demos}, we show visualizations of the rollout distributions and composition of demonstrations that are kept and filtered by \ours for the Strawberry and Sharpie tasks, along with the rollout distributions for the policy after filtering with \ours. 
We see that the rollout distributions at different checkpoints differ slightly, which we can leverage for cross-validation.
As shown by the red in the visualization of rollouts, some strategies lead to very low success rate when imitated by the learned policy. 
For the strawberry task, we find that \ours tends to filter out demonstrations where the robot approaches the strawberry from above instead of from the side, and for the sharpie task, \ours filters out demonstrations where the robot grasps the sharpie by the edge of the cap.
These strategies are unreliable for the robot to replicate, as evidenced by the low policy success with that strategy, but may be difficult for human demonstrators or annotators to recognize. 


\subsection{Variations of \ours}
\label{sec:exps-ablations}

In the following, we study particular design choices of \ours to understand their impact on policy performance in the simulated Robomimic square peg domain.

\paragraph{Importance of cross-validation using multiple checkpoints}

One key technical challenge is that the rollout distribution may not perfectly match the original demo distribution, and the classifier may overfit to classifying success vs failures on the rollouts it is trained on and fail to generalize to identifying unreliable strategies in the demonstrations.
Rollouts from different checkpoints will have slightly different distributions, so we use cross-validation to ensure that the chosen classifier will generalize well to different sets of rollout distributions and therefore better generalize to the demo distribution.
Many strategies for choosing these checkpoints may work--instead of taking evenly-spaced checkpoints, another strategy for choosing checkpoints is to use checkpoints near the initial plateau in policy performance where the policy first learns to fit the demos, and cross-validating on rollouts from the best checkpoint (which will occur later in the training run).
We find that this strategy is also effective for obtaining a classifier that generalizes well to the original demo distribution, as shown in Table~\ref{tab:ablations}.
Both of these strategies work because the rollout distributions used for training and validating the classifier will be representative of the demonstration data but will vary, so the chosen classifier will not overfit to rollouts produced by a single policy checkpoint. 
We find in Table~\ref{tab:ablations} that without cross-validation on sets of rollouts from different checkpoints, \ours performs poorly, as the classifier struggles to generalize to the demo distribution. 

\paragraph{Number of rollouts needed to train an effective classifier}

\begin{table}[t]
\centering
\resizebox{\columnwidth}{!}{
\begin{tabular}{lccccc}
\toprule
                                 & Base Policy & Training Loss & RCP   & Auto-IL & Demo-SCORE \\ \midrule
Square peg   (100 rollouts)                    & 79.9 & 79.2 & 87.1 & 88.2 & \textbf{94.4} \\ \midrule
Square peg (50 rollouts)         & 79.9 & 79.2 & 85.2 & 83.0 & \textbf{92.6} \\
Square peg (25 rollouts)         & 79.9 & 79.2 & 83.8 & 83.3 & \textbf{93.2} \\
Square peg (10 rollouts)         & 79.9 & 79.2 & 81.2 & 81.6 & \textbf{92.0} \\  \midrule \midrule
Peg insertion (100 rollouts) & 58.9 & 71.3 & 61.3 & 79.9 & \textbf{90.8} \\ \midrule
Peg insertion (50 rollouts) & 58.9 & 71.3 & 59.6 & 74.4 & \textbf{88.0} \\ 
Peg insertion (25 rollouts) & 58.9 & 71.3 & 56.3 & 70.2 & \textbf{89.1} \\ 
Peg insertion (10 rollouts) & 58.9 & 71.3 & 58.9 & 67.8 & \textbf{87.7} \\ 
\bottomrule
\end{tabular}
}
\caption{
\small
\textbf{Ablation on number of rollouts needed.} On two simulated tasks, \ours achieves strong performance using as few as 10 rollouts per checkpoint for classifier training. }
\label{tab:failure-rate}
\vspace{-4mm}
\end{table}

In this ablation, we test whether using fewer rollouts, such as 50 or even 10, from the first $C-1$ checkpoints to train the  classifier, and only 50 rollouts from the last checkpoint for the validation set, leads to the similar performance for \ours.  
This would be especially appealing for real-world settings where collecting rollouts may be expensive or time-consuming.
In Table~\ref{tab:failure-rate}, we find that using fewer rollouts leads to only a small decrease in performance, suggesting that only a small number of rollouts are needed to train an effective classifier. We note that other methods that leverage online rollouts have a much larger decrease in performance with fewer rollouts. Notably, with 10 rollouts per checkpoint and fewer than 100 rollouts total, \ours achieves 92\% average success rate across data mixtures, over 10\% higher than the next best method, Auto-IL, on the square peg task. Similarly, with only 10 rollouts per training checkpoint and 80 total rollouts, \ours achieves an 87.7 \% success rate, compared to a 58.9\% success rate for the base policy.

\paragraph{Using a trajectory classifier instead of a single-step classifier}

There are several ways to design the classifier. In our main experiments, we use a single-step classifier that takes in the state as input and outputs a single binary label, and we average the output across all states in the episode to determine whether to keep the demonstration.
Here we try using a trajectory classifier (a small transformer) that takes in the entire trajectory as input and outputs a single binary label for the entire trajectory.
Because the length of trajectory is more directly correlated with the success of the trajectory than the strategy, we only use the first 100 steps of the trajectory as input to the classifier.
In Table~\ref{tab:ablations}, we find that the trajectory classifier is as effective as the single-step classifier on the simulated square peg task, even giving slightly higher average performance, showing that multiple types of classifiers may be appropriate for \ours.

\paragraph{Filtering at the chunk level instead of episode level}

In our main experiments, we filter at the episode level, keeping or discarding the entire demonstration based on the classifier output.
However, \ours can also be applied at the chunk level, where we keep or discard individual chunks of the demonstrations based on the classifier output.
In Table~\ref{tab:ablations}, we find that filtering at the chunk level is also very effective in our simulated experiments, matching the performance of episode filtering, showing that \ours can be applied reliably at different levels of granularity. This is especially appealing for long-horizon tasks, to not discard valuable information in segments of demonstrations.


\paragraph{Importance of classifier regularization}
Finally, we ablate the effect of classifier regularization on the performance of \ours. 
We test the effect of setting weight decay to 1e-4 and removing dropout on the classifier, and find that the performance of \ours decreases slightly but is still comparable when these are removed, as seen in Table~\ref{tab:ablations}. Thus, the classifier is robust even without heavy regularization, due to cross-validation. 

\section{Limitations}

In this paper, we propose \ours, which curates demonstration datasets by pruning suboptimal examples estimated from online robot experience. Our experiments on a variety of simulated and real-world tasks show that \ours significantly improves the robustness of learned robot policies compared to training on all available demonstrations and to prior methods that leverage online rollouts.
While we are excited about the potential of \ours, there are several limitations and directions for future work that we discuss as follows.
First, \ours relies on the assumption that the policy rollouts are representative of the original demonstration distribution, which may not always be the case if the policy is incapable of modeling the full distribution of demonstrated behaviors while avoiding mode collapse.
In particular, the policy may not be able to replicate all strategies shown in the demonstrations, and so the policy rollouts may not perfectly match the original demo distribution.
Also, \ours may not be effective in settings where the demonstrations are all so low quality or sparse that the initial policy is unable to successfully complete the task at all, although we find that we can sometimes address this using subtask scores, as done in the Jellybeans task.
Lastly, \ours may reduce the coverage of states for which the policy is in distribution. As our goal is instead to maximize final performance on a given task, we no longer aim to maximize state coverage in demonstration curation and instead aim to maximize the reliability of demonstration strategy. Thus, \ours is helpful in maximizing final performance, so we recommend using it for post-training and not for pre-training stages where training on wider distributions may be helpful. 
Our results highlight the importance of high-quality demonstrations and suggest that careful curation can significantly enhance the final performance of learned robot policies.


\section*{Acknowledgments}
We thank Suvir Mirchandani, Jonathan Yang, Zach Witzel, Kaylee Burns, and others in the Stanford IRIS lab for support with real-robot experiments. This work was supported in part by an NSF CAREER award and ONR grants N00014-21-1-2685 and N00014-22-1-2621. A.S. Chen acknowledges support by the NSF GRFP and an OpenAI superalignment grant. Y. Liu acknowledges support by an SNSF Postdoc fellowship.

\bibliographystyle{plainnat}
\bibliography{example, bibtex/robotics, bibtex/social, bibtex/data, bibtex/rl}  

\newpage
 \appendix
 \section{Appendix}

 \begin{table*}[h]
\centering
{
\begin{tabular}{@{}p{4.5cm}p{1.5cm}p{2.0cm}p{1.5cm}p{1.5cm}p{1.5cm}p{1.5cm}@{}}
\toprule
                                                       & 100 Human, 100 SquareA & {100 Human, 50/50 SquareA/B} & {200 Human, 20 SquareA} & {200 Human, 100 SquareA} & {200 Human, 200 SquareA} & {200 Human, 400 SquareA} \\ 
\toprule
Num Demos Filtered Out           & 104         & 93            & 35            & 142           & 200           & 400           \\
\bottomrule
\end{tabular}
}
\caption{
        \small 
        \textbf{Number of demonstrations filtered by \ours.} \ours is effective even with strongly lopsided heterogeneous mixtures and adjusts accordingly depending on how much should be filtered from a given dataset. 
    }
    \label{tab:demos-filtered}
\end{table*}

\begin{table*}[h]
\centering
\resizebox{0.85\textwidth}{!}{
\begin{tabular}{l|c|cccccc}
\toprule
Classifier MLP Size & Base Policy & {[}8, 8{]} & {[}8, 8, 8{]} & {[}16, 16{]} & {[}16, 16, 16{]} & {[}32, 32{]} & {[}32, 32, 32{]} \\
\toprule
Success Rate    & 79.9        & 94.4       & 95.8          & 95.3         & 94.4             & 94.0         & 95.1            \\
\bottomrule
\end{tabular}
}
\caption{
        \small 
        \textbf{Ablation on Classifier Size}. For the square task, we report the average percent success rate for the across data splits when using \ours with step MLP classifiers of different sizes. For each MLP architecture, we report the number of hidden units in each hidden MLP layer and the average success rate of the final Demo-SCORE policies. The main experiments used an ``[8, 8]" architecture with two hidden layers with 8 units each.
    }
    \label{tab:cls-size}
\end{table*}

\subsection{Training Details}

\paragraph{Simulated ALOHA domain with ACT policy}
 
We train every sim ACT policy for 300,000 steps and measure policy success rate with 256 rollouts every 10,000 steps. The reported success rate is the maximum success rate across each of these evaluations. Other than extending training to 300,000 steps, we use the default policy hyperparameters for the LeRobot codebase \cite{cadene2024lerobot} (which are based on the hyperparameter choices in the original ACT paper).
We use precisely the same policy training procedure for initial policies, our method, and comparisons.

\paragraph{Robosuite Domain with Diffusion Policy}

We train all policies for 1,000 epochs.
Each evaluation is done with 256 rollouts. The success rate reported for each policy is the maximum success rate in any evaluation.  All policy hyperparameters are the same as those in the original Diffusion Policy paper.

\subsection{Additional Method Details}

\paragraph{\ours Checkpoint Selection}

For our method, we collect rollouts for use by Demo-SCORE or the baselines using rollouts from epochs 250, 500, 750, and the last checkpoint. For the ALOHA Sim experiments, checkpoints at steps 70,000, 150,000, 220,000, and 300,000 were used to collect rollouts. For the ablation with \ours classifier training checkpoints chosen to be at the initial plateau in policy performance, those checkpoints are chosen from the same set of checkpoints used for policy evaluation.

For real-world experiments, upon training the initial policy past the optimal range, one or two rollouts were taken for a wide variety of checkpoints to determine the approximate range of checkpoints where the policy performed best. Within that range, 5 evenly spaced epochs were used for evaluation and rollout collection for each policy trained on that task. Checkpoints with poor performance (success rates less than half that of the best checkpoint’s) were discarded.

\paragraph{\ours Classifier Training Details}

There are $C - 1$ training datasets $D_{\pi_{0,i}}$ for $i = 1, 2, …, C - 1$, and we use $C = 4$ for the simulated experiments and $C=5$ for real-world experiments. For each of these, 
a classifier was trained. For each classifier training run, the best checkpoint was chosen using our cross-validation metric (the unweighted binary-cross entropy loss of our classifier on the cross-validation set, $D_{\pi_{0,C}}$). Then, of the classifier runs, the best was chosen with the same metric. When training step classifiers, the loss was computed for each step in the dataset, and then averaged. For the trajectory classifier variation, the loss was computed across trajectories in the dataset and then averaged.
For the ablations without cross-validation, a total of 200 rollouts from the same policy checkpoint (the best performing one) were used. For each training run, these were randomly divided into a set of 100 train and 100 val rollouts. 
All classifiers use dropout of 0.3 and AdamW weight decay of 0.1.

\subsection{Additional details on comparisons}

\paragraph{Training Loss method description}
To compute the training loss used in the training loss weighting baseline, we compute the loss using the policy from the initial training run using all demos (the best checkpoint) for every example drawn from a given episode in the set of original demos. We then take the mean across all examples for a given episode to get a mean training loss for that episode. For diffusion policy, we do this eight times for each episode and take the mean due to stochasticity in the diffusion policy forward pass. We then associate a weight with each episode in the demo set by taking the inverse of the mean loss for that episode. We normalize this set of weights by subtracting the minimum value and dividing by the standard deviation. Having calculated these episode-wise weights, we train the policy again, multiplying the loss for each example by the weight associated with the episode that example came from.

\subsection{Additional Simulated Result Tables}

We provide the number of demonstrations filtered out by \ours in Table~\ref{tab:demos-filtered} for the square peg task, showing that \ours adjusts the amount of demos filtered depending on the quality mixture of the original demonstrations. In Table, \ref{tab:cls-size}, we show that the performance of DemoSCORE on the square task is consistent as we vary the number of hidden units and layers in the step classifier MLP.

\subsection{Additional Details on Jellybean Task}

The subtasks of the jellybean task are:
\begin{itemize}
    \item retrieving the cup from the cup dispenser
    \item placing the cup on the table next to the jellybeans dispenser
    \item pushing the cup underneath the spout of the jellybean dispenser
    \item filling the cup with jellybeans
    \item moving the filled cup to the front of the table
\end{itemize}
We provide success rates by subtask for the Jellybean task in Table \ref{tab:jelly-subtask}. Additionally, we define an aggreated subtask score as the fraction of subtasks completed in an episode. For example, if the robot completed the first 3 of 5 subtasks in a given episode, that episode would get a score of 60\%. As the base policy trained on all demonstrations did not ever succeed on the full task, this aggregated subtask score was used as the target label to supervise classifier training for this task.

\subsection{Impact of Demo-SCORE Filtering on OOD Performance} \label{ssec:ood}

One potential concern is that Demo-SCORE narrows the state distribution in the original dataset, which may affect generalization capabilities. To understand this, we evaluate two
generalization settings in the square peg domain: (1) OOD initial conditions, and (2) different
train-test splits across in-task variations of different initial
conditions. 

For (1), in Table \ref{tab:ood-evals}, we show the success rate of the square task policies trained on each data split within a modified environment where the xy dimensions of the region where the square nut is initialized are expanded by 50\% (easy) and 100\% (hard) beyond the standard environment. As the demos used to train these policies were all from the original environment, the nut's initial position is OOD in these expanded regions relative to the datasets used to train these policies. We find that Demo-SCORE (with either full episode filtering or chunk-level filtering) does not harm OOD generalization in this setting compared to the base policy or Auto-IL and even improves performance whe the environment is initialized in these OOD states.

For (2), our goal is to test the  test the policy’s generalization ability to initial conditions that
are both in-distribution and OOD but have mostly or only suboptimal data in some regions. Specifically, we train new policies on new datasets in which there is a correlation between the initial state of the nut and the strategy used in each demo. We divide the initial state space of the nut by the orientation of the nut, defining "even" and "odd" regions corresponding to the quadrant in which the angle of rotation around the z-axis of the nut falls. (In other words, if the angle by which the nut is rotated around the z-axis relative to its reference position falls between 0 and 90 degrees or 180 and 270 degrees, it is considered within the "odd" quadrant region).  

In these new datasets, we control how many Human and Square A demos there are in which the nut's initial orientation is Odd or Even. Each dataset has 200 total demonstrations, 100 for the odd and even region.  In the Even region, we vary the percentage of demos that are on the Square A mode from 100\% to 50\%, in increments of 10\%. We either chose the opposite ratio of modes in the Odd region or keep 50 demos from each modes in the odd region. We present the success rates of the policies trained on these datasets in Table \ref{tab:cor-eval}.

Demo-SCORE improves performance when different ratios of
high-quality and suboptimal demonstrations exist for different
initial conditions and does not degrade performance
for states dominated by suboptimal demonstrations, suggesting
that Demo-SCORE does not filter out suboptimal data that
contributes to generalization success. However, in some of the most extreme cases (such as one data split where 100\% of demos in the even region of initial state space have the Square A strategy), \ours can slightly under-perform other methods. Nonetheless, these results suggest that Demo-SCORE generally does not excessively narrow the state distribution in a way that harms policy
generalization. Intuitively, Demo-SCORE aims to keep any
data that contributes to policy success and therefore may retain
suboptimal strategies if they appear in successful rollouts.
Additionally, if there are settings where filtering may be overly
restrictive, this can be mitigated by common alternatives, e.g.
reweighting. Our key contribution is an approach for estimating
the quality of a demo or chunk, and this score can be applied
in various ways, including but not limited to filtering.

\begin{table*}[h]
\centering
\begin{tabular}{l|p{1cm}p{1cm}p{1cm}p{1cm}|p{1cm}p{1cm}p{1cm}p{1cm}}
\toprule
                           & \multicolumn{4}{c}{50\% Expanded XY}                                                                                                           & \multicolumn{4}{c}{100\% Expanded XY}                                                                                                          \\
Data Split                 & Base Policy & Auto-IL & Demo-SCORE & Demo-SCORE (chunk) & Base Policy & Auto-IL & Demo-SCORE & Demo-SCORE (chunk) \\ \midrule
100 Human 100 SquareA     & 65.2                            & 71.9                               & 85.9                           & 85.9                                   & 51.6                            & 63.3                               & 73                             & 72.3                                   \\
100 Human, 50/50 Square A/B & 66.4                            & 77.7                               & 88.7                           & 85.2                                   & 54.3                            & 64.8                               & 74.2                           & 72.3                                   \\
200 Human, 20 Square A     & 89.8                            & 90.6                               & 92.6                           & 92.2                                   & 80.1                            & 80.9                               & 83.2                           & 82                                     \\
200 Human, 100 Square A    & 73.4                            & 84.8                               & 89.1                           & 90.6                                   & 64.5                            & 75.4                               & 81.3                           & 82                                     \\
200 Human, 200 Square A    & 75.4                            & 87.1                               & 93                             & 94.5                                   & 61.7                            & 82.4                               & 83.4                           & 80.5                                   \\
200 Human, 400 Square A    & 68                              & 76.2                               & 94.1                           & 94.9                                   & 54.7                            & 65.6                               & 83.6                           & 84.4                                   \\ \midrule
Average                    & 73.0                            & 81.4                               & 90.6                           & 90.6                                   & 61.2                            & 72.2                               & 79.8                           & 78.9                                 \\ \bottomrule
\end{tabular}
\caption{
        \small 
        \textbf{OOD Evaluation of Square Task Policies}. The success rates of the base policies, policies trained with Auto-IL, and two variants of \ours, on OOD environments. The xy dimensions of the region in which the square nut's initial position is sampled (uniformly) are expanded by 50\% and 100\% relative to the original environment. The demonstrations used to train the evaluated policies come from the original environment, and these policies are thus OOD on the modified environments. \ours outperforms the base policy and the Auto-IL policy when the environment is initialized to OOD states. 
    }
    \label{tab:ood-evals}
\end{table*}

\begin{table*}[h]
\centering
\begin{tabular}{l|p{1cm}p{1cm}p{1cm}|p{1cm}p{1cm}p{1cm}|p{1cm}p{1cm}p{1cm}}
\toprule
                              & \multicolumn{3}{c}{Standard Environment}                                                               & \multicolumn{3}{c}{50\% Expanded XY}                                                         & \multicolumn{3}{c}{100\% Expanded XY}                                                        \\
Data Mixture                  & Base Policy & Auto-IL & Demo-SCORE & Base Policy & Auto-IL & Demo-SCORE & Base Policy & Auto-IL & Demo-SCORE \\ \midrule
100\% SqA Even, 100\% Human Odd   & 54.3                            & 62.9                        & 69.9                           & 57.0                              & 56.2                        & 63.7                           & 46.9                            & 49.6                        & 51.2                           \\
90\% SqA Even,  90\% Human Odd  & 59.8                            & 63.3                        & 78.1                           & 57.4                            & 60.5                        & 71.5                           & 50.4                            & 52.3                        & 62.9                           \\
80\% SqA Even,  80\% Human Odd  & 66.4                            & 73.4                        & 86.7                           & 62.9                            & 68.4                        & 80.9                           & 53.9                            & 56.6                        & 68.4                           \\
70\% SqA Even,  70\% Human Odd  & 69.5                            & 83.6                        & 91.4                           & 68.8                            & 75.4                        & 89.8                           & 62.5                            & 60.5                        & 78.1                           \\
60\% SqA Even,  60\% Human Odd  & 67.2                            & 78.1                        & 88.7                           & 64.8                            & 77.0                         & 86.3                           & 57.0                              & 67.6                        & 70.7                           \\
50\% SqA Even,  50\% Human Odd  & 67.6                            & 76.2                        & 91.4                           & 65.2                            & 71.9                        & 85.5                           & 57.0                              & 62.1                        & 75.4                           \\ \midrule
100\% SqA Even,  50\% Human Odd & 45.7                            & 56.2                        & 52.3                           & 43.8                            & 49.6                        & 51.2                           & 43.8                            & 40.6                        & 39.8                           \\
90\% SqA Even,  50\% Human Odd  & 48.8                            & 60.5                        & 69.9                           & 44.5                            & 61.3                        & 65.6                           & 41.8                            & 48.4                        & 52.0                             \\
80\% SqA Even,  50\% Human Odd  & 61.7                            & 68.0                          & 85.2                           & 60.9                            & 64.8                        & 79.7                           & 53.1                            & 53.9                        & 66.4                           \\
70\% SqA Even,  50\% Human Odd  & 63.3                            & 69.5                        & 78.1                           & 59.8                            & 64.5                        & 75.8                           & 53.1                            & 58.6                        & 64.1                           \\
60\% SqA Even,  50\% Human Odd  & 67.6                            & 76.2                        & 89.8                           & 65.2                            & 77.0                          & 86.3                           & 59.8                            & 69.1                        & 71.1            \\ \bottomrule              
\end{tabular}
    \caption{
        \small 
        \textbf{Success Rates on Correlated Datasets}. The success rates on the square task of base policies, Auto-IL, and \ours policies when the original data mixture has a correlation between the initial orientation of the square nut and the strategy exhibited in each demo as described in section \ref{ssec:ood}. Each row corresponds to an initial data mixture and is denoted by the percentage of demos within the even region that are employ the Square A strategy and percentage of demos within the odd region that are Human demos (which employ the same strategy). We present evaluations for the standard environment, as well as OOD evaluations on a modified environments where the square nut's initial xy position is initialized within a region expanded 50 or 100\% relative the standard environment. \ours generally outperforms the base policy and Auto-IL even when there is a correlation between initial state and demonstrated strategy \textit{and} the evaluation environment is OOD. 
    }
    \label{tab:cor-eval}
\end{table*}


\end{document}